\documentclass{article}
\usepackage[preprint]{spconf}
\usepackage{amsmath}
\usepackage{amssymb}
\makeatletter
\def\maketag@@@#1{\hbox{\m@th\normalfont\normalsize#1}}
\makeatother
\usepackage{graphicx}
\usepackage{subcaption}
\usepackage{relsize}
\usepackage{cleveref}
\crefname{figure}{Fig.}{Figs}
\usepackage[%
square,%
numbers%
]{natbib}
\usepackage[usenames,dvipsnames]{xcolor}
\usepackage{tikz}
\usetikzlibrary{positioning}
\usetikzlibrary{calc}
\usetikzlibrary{fit}
\usepackage{cuted}
\usepackage{enumitem}
\usepackage{calc}
\usepackage{textcomp}

\definecolor{logoH1}{RGB}{51,143,152}
\definecolor{logoH2}{RGB}{46,169,72}
\definecolor{logoM}{RGB}{76,185,56}
\definecolor{logoI}{RGB}{125,201,44}

\tikzstyle{imageNode}=[inner sep=0,anchor=south west]
\tikzstyle{myGrid}=[help lines, xstep=3.0, ystep=0.3, white, line width=1pt]
\tikzstyle{labelNode}=[]
\tikzstyle{waveIndicator}=[draw,ultra thick,color=red,->]

\pdfoutput=1
\title{\uppercase{Robust Registration of Calcium Images by Learned Contrast Synthesis}}
\name{ John A.~Bogovic, Philipp Hanslovsky, Allan Wong, Stephan Saalfeld}
\address{HHMI Janelia Research Campus\\
  19700 Helix Drive, Ashburn, VA 20147}

\begin{document}

\maketitle

\begin{abstract}
    Multi-modal image registration is a challenging task that is vital
    to fuse complementary signals for subsequent analyses.  Despite much research into
    cost functions addressing this challenge, there exist cases in which 
    these are ineffective.  In this work, we show that (1) this is true
    for the registration of in-vivo \emph{Drosophila} brain volumes
    visualizing genetically encoded calcium indicators to an nc82 atlas and (2) that 
    machine learning based contrast synthesis can yield improvements.
    More specifically, the number of subjects for which the registration
    outright failed was greatly reduced (from 40\% to 15\%) by using a
    synthesized image.
    
\end{abstract}

\section{Introduction}
\label{sec:intro}
Two photon confocal microscopy with genetically encoded calcium indicators
has been used successfully for the monitoring of neural activity in
defined neuronal subtypes, for instance, in the \emph{Drosophila}
antennal lobe\cite{Wang2003}.  To monitor larger, more heterogeneous
populations of neurons in the whole brain, it is both necessary and challenging to identify the
activated neuron reliably.
Neuron identification within a large pool of references greatly benefits from registration to the reference \citep{Costa15}.
Here, we use the fruit fly
\emph{Drosophila melano\-gaster} as our model organism, and  experiment
with different approaches to align baseline fluorescence images of
experimental brains to a bridging template brain with the same driver
line and a neuropil marker (see section 2.1).  The neuropil
marker is then used to align the template brain to a database of
individually segmented neurons. 

While both the template and experimental images measure similar
quantities, they are imaged with different modalities and thus have
differing image content, noise properties, and dynamic range.
Multi-modal image registration is a common and challenging task 
in medical and biological imaging and an active area of research.

\begin{figure}
\centering%
\setlength\tabcolsep{1mm}%
\begin{tabular}{@{}cc@{}}
\includegraphics[width=\columnwidth/2-2\tabcolsep]{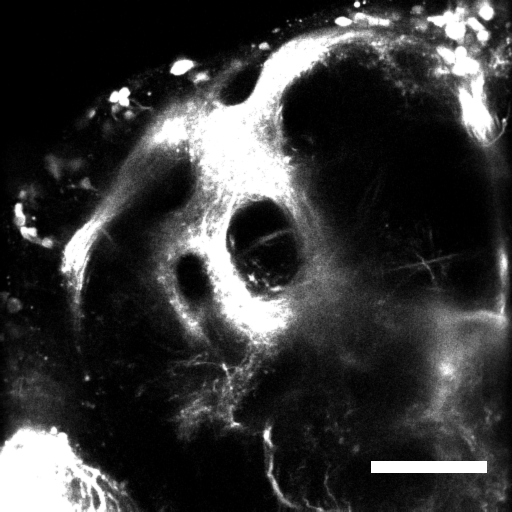} &
\includegraphics[width=\columnwidth/2-2\tabcolsep]{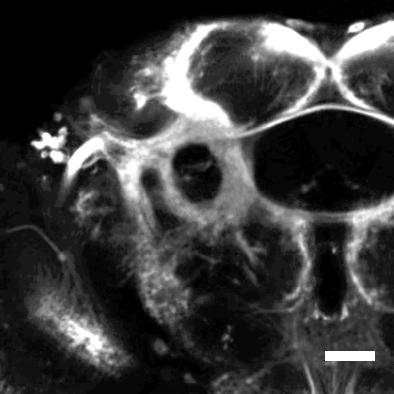} \\
    (a) subject image ($xy$) & (b) template ($xy$) \\ [1ex]
    \includegraphics[width=\columnwidth/2-2\tabcolsep,height=0.489847716\columnwidth/2-0.979695431\tabcolsep]{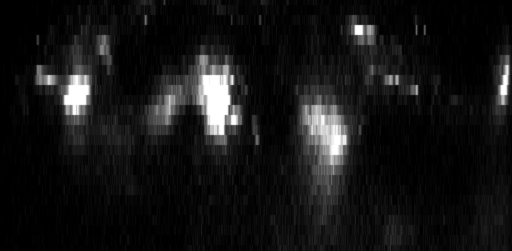} &
    \includegraphics[width=\columnwidth/2-2\tabcolsep,height=0.489847716\columnwidth/2-0.979695431\tabcolsep]{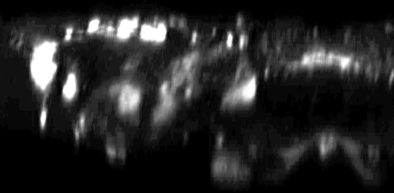} \\
    (c) subject image ($xz$) & (d) template ($xz$)
    
\end{tabular}
\caption{Example subject (a,c) and template (b,d) image.  Notice the
low $z$-resolution in the subject (c) compared to the approximately isotropic
template (d).  Scale bars 50\;\textmu{}m.}
\label{fig:imageExamples}
\end{figure}

\subsection{Related work}
\label{sec:related}

Kybic et al.~\cite{Kybic2014} simultaneously segmented and registered
histological images.  The registration step minimizes the mutual
information of class labels.  This approach is an excellent choice
when the modalities have corresponding, but differently appearing pixel
classes but may have difficulties in cases where the boundaries of
various labels are unclear, or when image content differs significantly.  

A different approach appears in Roy et al.~\cite{Roy2013} where 
the target modality is ``synthesized'' from the source modality
directly.  Their proposed method uses a registered pair (one from each
modality) of images of the same subject as an ``atlas.''  Their method
uses a patch-based search with heuristics designed for MRI to estimate
the target modality from source. They show that intra-modality
registration using the result outperforms inter-modality methods.

In their survey, Sotiras et al.~\cite{Sotiras2013} describe many other
alternative approaches.  Most related to our method are the approaches by Wein et al.~\cite{Wein2008} who simulate
an ultrasound image from CT using imaging physics and known tissue
properties, and Michel et al.~\cite{Michel2010} who use a mixture of experts
and MRF to learn the probability of a target intensity conditioned on a
source image patch.

\section{Methods}
\label{sec:methods}

Our method is inspired by other inter-modality registration methods that
``synthesize'' a target image modality from a source image modality.
Unlike much previous work, an ``atlas pair'' of images (one from each modality)
is not available for our application.  Instead, we manually register the two modalities
for some set of subjects to create a ``silver standard atlas.''  Given
the inherent error in this atlas, we opt to learn a mapping between 
the modalities via a classifier, similar to \cite{Michel2010}, rather than use patch correspondences 
directly, as in~\cite{Roy2013}.  Once this mapping is learned, it can 
be applied to the subject modality to produce an image more similar to
the target, with the hope that the resulting image can be more reliably
registered with standard algorithms than the original modality.

% TODO emphasize that we don't have ground truth in the sense that Roy
% did, we need to create it ourselves, with 'bigwarp' - done
\subsection{Imaging and Preprocessing}
\label{sec:imaging}
We express the calcium indicator GCaMP6s in Fruitless neurons and
collect volumetric images at 512\,\texttimes\,512\,\texttimes\,42 pixels
(at 0.43\,\texttimes\,0.43\,\texttimes\,5 \textmu{}m/pixel resolution).
Each volume was collected at 0.95\,Hz and 200 volumes were averaged to
create a baseline volume image with high signal to noise.  This serves
as the subject image.  The 4D volumes can be analyzed independently to
extract calcium signals.
 
Separately, we collected the bridging template brain of the same
Fruitless driver line crossed with myr::GFP, which uniformly labels the
membrane of the neuron. The brain was then immunohistochemically stained
with anti-GFP and with nc82 antibodies to label the neuropils, cleared
with DPX and imaged with a Zeiss confocal using a 20\texttimes{} air
objective.  The nc82 channel was then used to register the Fru-myr::GFP
sample to a common reference.  This bridging template brain, after
warping to the common reference, has 1\,\texttimes\,1\,\texttimes\,1
\textmu{}m/pixel resolution.

We preprocess the images by clipping intensities above the
99\textsuperscript{th} percentile, followed by Gaussian smoothing with a
1.5\,\texttimes\,1.5\,\texttimes\,1.5\,\textmu{}m kernel, and finally
linearly scaling the intensities to the range [0, 255].  This was
applied to both the template and subject images.

% This section might do better in experiments/results section
\subsection{Baseline Registration}
\label{sec:baseline}
We attempt to register the subject images to the template using the SyN
algorithm~\cite{Avants2009} (part of the
ANTs\footnote{http://stnava.github.io/ANTs} toolbox) using several cost
functions: sum of squared differences (SSD), normalized cross correlation (CC), and mutual information (MI).  The preprocessed template and
subject images were used as inputs to SyN.

% Getting the ground truth
\subsection{Ground Truth}
\label{sec:groundTruth}
We first manually coregistered several subject image volumes to the
template by placing corresponding landmarks and using a thin plate
spline transformation to generate a dense warping.  This was important
because a set of registered images is needed in order to learn a mapping
between the modalities, as will be explained below in
section 2.4.

The registration task is quite challenging even for human annotators.
One human annotator (JB) placed landmarks on all images to produce a
registration.  Another annotator (SS) began with the first annotator's
landmarks on the target images only, and independently placed landmarks
on the moving image.  Using these measurements, we
can determine whether a given registration algorithm performs the task
with ``human-like'' precision.

To facilitate manual registration with landmark placement, we developed
\emph{BigWarp}\footnote{https://github.com/saalfeldlab/bigwarp}, a tool for fast,
interactive, and deformable alignment of large 2D and 3D images.  It builds upon \emph{BigDataViewer}
\citep{Pietzsch2015} for visualization and data sources which enables
rapid navigation of very large images with arbitrary 3D reslicing and zoom.
We made BigWarp publicly available through the ImageJ distribution
Fiji \citep{Schindelin2012}\footnote{https://fiji.sc/bigwarp} contributing real time deformable 2D and 3D transformation for almost arbitrarily large images.

\subsection{Learning the synthesis mapping}
\label{sec:mapping}

\begin{figure}[t]
\centering
\begin{tabular}{ccc} 
    \rotatebox{90}{ \parbox{1.2in}{\centering Fly $1$ (training)}} &
    \includegraphics[height=1.2in]{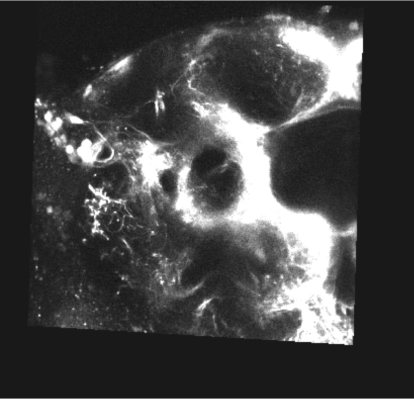} &
    \includegraphics[height=1.2in]{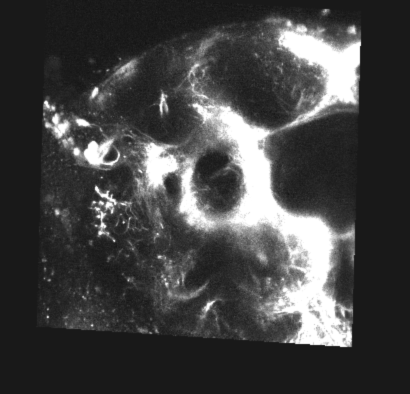} \\
    \rotatebox{90}{ \parbox{1.2in}{\centering Fly $2$}} &
    \includegraphics[height=1.2in]{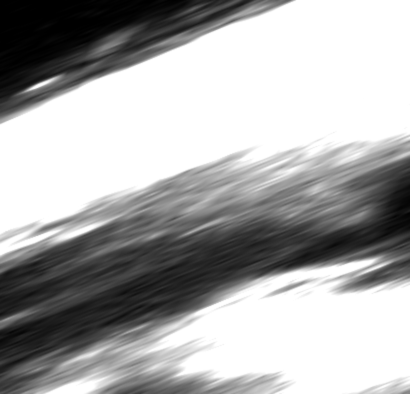} &
    \includegraphics[height=1.2in]{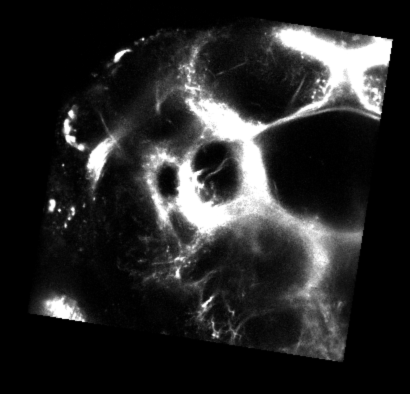} \\
    \rotatebox{90}{ \parbox{1.2in}{\centering Fly $3$}} &
    \includegraphics[height=1.2in]{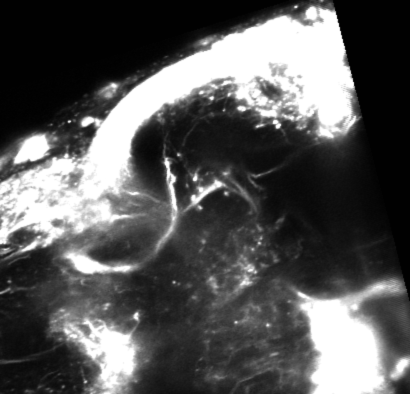} &
    \includegraphics[height=1.2in]{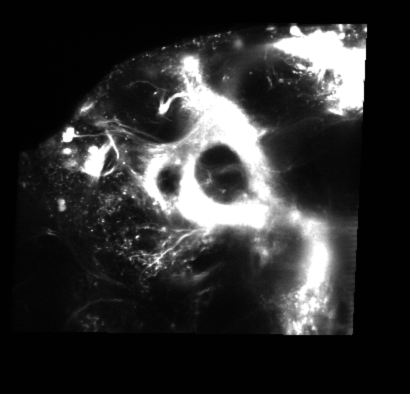} \\
    \rotatebox{90}{ \parbox{1.2in}{\centering Fly $4$}} &
    \includegraphics[height=1.2in]{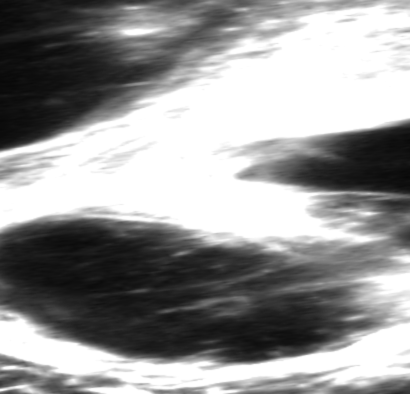} &
    \includegraphics[height=1.2in]{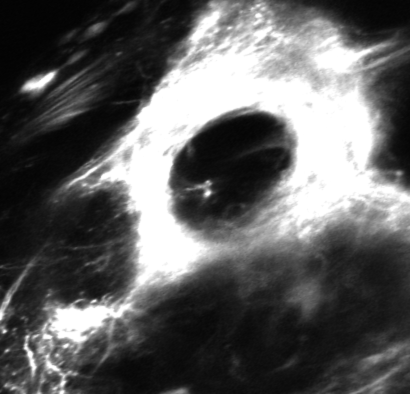} \\
     & Baseline & Synthesis \\
\end{tabular}
\caption{ Registration results for baseline (left) and random forest
    synthesis (right).  Both methods succeed for fly 1 (one of the
    training subjects), only the synthesis succeeds for flies 2 and 3, and both
    methods fail for fly 4. } 
\label{fig:resultsImage}
\end{figure}

We used manually registered subject + template image pairs to learn a mapping (``synthesis'') from the subject
intensities to the template intensities.  This analysis was done in
subject image space in order to avoid the significant upsampling that
would be necessary to transform the lower-resolution subject image into
the template image space.

To learn the mapping from source to target modalities, we used boosted
decision trees (BDT) \citep{Becker2013a} and a random
forest (RF) classifier (using the implementation in VIGRA \citep{Kothe1999}).

Boosted decision trees were trained using a squared loss for $10$k
iterations, with a shrinkage factor of $0.01$, a subsampling factor of
$0.2$, to a max depth of $3$.  The random forests were trained to purity
with $1000$ trees and a subsampling factor of $0.1$.  We ran experiments
using two kinds of features for both algorithms: (a)
pixel intensities inside a 5\,\texttimes\,5 \texttimes\,3 patch
around the pixel of interest, (b) multi-scale
gradient, intensity, and texture based features.  For both methods, $200$k samples were randomly
drawn from two training subjects.  Surprisingly, the resulting
registrations using (b) had larger errors across all subjects compared to using
(a).  The classifier does not
predict the continuous pixel intensity of the template image, but one 
of 10 classes derived from binning the pixel intensities by every
10\textsuperscript{th} percentile.  We experimented with other schemes
as well; this had the best performance of those we tried.

Once a classifier is learned, we apply it to all pixels of the subject
image to obtain the ``synthetic'' template image.  This is registered
with SyN, with identical parameters as used for the baseline
registration (see section 2.2).

\section{Experiments}
\label{sec:exps}

\begin{figure*}
    \includegraphics[width=\textwidth]{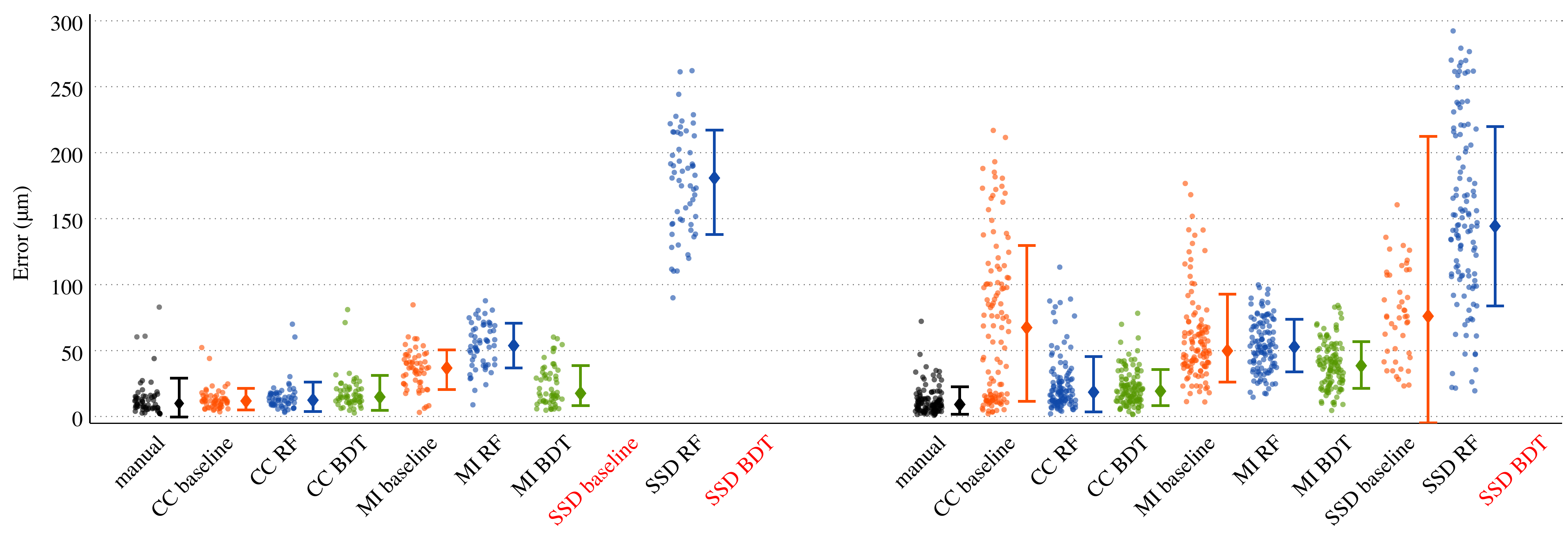}
\caption{ A scatter plot showing landmark errors for different registration
    methods for training (left group) and testing subjects. In each
    group, the left most points (black) show inter-human landmark error. Next are
    baseline (orange), random forest (RF) (blue), and boosted decision tree
    (BDT) (green). Results for each of these three methods is given with
    three cost functions in ANTS, cross correlation (CC), mutual
    information (MI), and sum of squared differences (SSD).
    The landmark errors are plotted as points, the median as a diamond,
    and the bar gives the mean $\pm$ one standard deviation.  There are 
    $72$ and $180$ landmarks for the training and testing subjects,
    respectively.
}
\label{fig:errplot}
\end{figure*}

We inspected each registration result using CC visually and 4 of 7
baseline results are (qualitatively) useful, where 6 of 7 synthesis
results are useful.  F shows four of these results 
for baseline (left) and synthesized (RF) experiments for three different
subjects, all using ANTS with a cross correlation cost. 

We computed errors (euclidean distance) between landmarks placed by a
human annotator, and landmarks transformed by a registration.
Fig.~2 shows landmark errors for various experimental
images and registration techniques.  The left group of plots shows
errors for the two subjects used to train the contrast synthesis
mapping; the right group shows errors for the four subjects left out for
testing.  The leftmost scatter group (black) plots the inter-human
landmark placement error.  The next nine plots show errors for the
``baseline'' registration method (the SyN algorithm on smoothed image
data), and the proposed synthesis approach for two different
classifiers, each with three choices of cost function.  We note that for
some subjects, the SSD cost resulted in a transformation with a singular
affine part which is equivalent to total failure.  We indicated those by a red $x$-axis label.

The pattern shown by example in Fig.~2 is consistent with and explains the distribution of errors
in Fig.~3.  Many errors are small (on the
order of inter-human error).  Catastrophic landmark transfer errors are more common in the baseline than in any of the synthesized images.

\section{Discussion}
\label{sec:discussion}

A surprising outcome of our experiments is the extent to which mutual
information underperformed relative to cross correlation for
both baseline and synthesis experiments.  This could be due to
the rather small dynamic range of the acquired images.  We also observed
that the warp field estimated by SyN was often very small for subjects
that failed.  This seemed to follow a poor initial estimate of the
affine part of the transformation which may explain why an additional
warp could not further reduce the cost function.  

Another observation of note is the fact that our method was trained
only on subjects for which the baseline method succeeded.  Still, 
when applied to more challenging subjects (for which the baseline
failed) the classifier generalized sufficiently well to produce
a synthesized image that could be registered successfully to the
template.  Training on larger and more diverse set of subjects could
further improve robustness in the future.

We also note that unlike~\cite{Roy2013}, our proposed method does not
use heuristics to aid in the synthesis mapping.  As such, we hope that
this approach may be applicable to a wider variety of domains.  However,
a drawback is the fact that the synthesized images produced here do not 
resemble the target modality to a human (but evidently are more similar
to cost functions used in registration).  Exploring alternative models
or algorithms to implement the mapping (convolutional networks, or
mixture of experts, as in \cite{Michel2010}) are interesting avenues of future
research.

In conclusion, our results show that learning to synthesize the
template modality from the subject modality results in more robust
registration performance (i.e.~fewer subject registrations resulted in
unusable results).  Secondly, cross correlation outperforms other cost
function for the registration of these image data.

\section{Acknowledgments}

The authors would like to thank Barry Dickson and Kaiyu Wong for their
valuable feedback and discussions regarding image acquisition.

\small
\bibliographystyle{IEEEbib}
%\bibliography{references}

\begin{thebibliography}{10}

\bibitem{Wang2003}
J.~W. Wang, A.~M. Wong, J.~Flores, L.~B. Vosshall, and R.~Axel,
\newblock ``{Two-photon calcium imaging reveals an odor-evoked map of activity
  in the fly brain.},''
\newblock {\em Cell}, vol. 112, no. 2, pp. 271--82, 2003.

\bibitem{Costa15}
M.~Costa, J.~D. Manton, A.~D. Ostrovsky, S.~Prohaska, and G.~S. Jefferis,
\newblock ``Nblast: Rapid, sensitive comparison of neuronal structure and
  construction of neuron family databases,''
\newblock {\em bioRxiv}, 2015.

\bibitem{Kybic2014}
J.~Kybic and J.~Borovec,
\newblock ``{Automatic simultaneous segmentation and fast registration of
  histological images},''
\newblock in {\em Int. Symp. Biomed. Imag.}, 2014, pp. 774--777.

\bibitem{Roy2013}
S.~Roy, A.~Carass, and J.~L. Prince,
\newblock ``{Magnetic Resonance Image Example-Based Contrast Synthesis},''
\newblock vol. 32, no. 12, pp. 2348--2363, 2013.

\bibitem{Sotiras2013}
A.~Sotiras, C.~Davatzikos, and N.~Paragios,
\newblock ``{Deformable medical image registration: a survey},''
\newblock {\em IEEE Trans Med Imaging}, vol. 32, no. 7, pp. 1153--1190, 2013.

\bibitem{Wein2008}
W.~Wein, S.~Brunke, A.~Khamene, M.~R. Callstrom, and N.~Navab,
\newblock ``{Automatic CT-ultrasound registration for diagnostic imaging and
  image-guided intervention},''
\newblock {\em Medical Image Analysis}, vol. 12, no. 5, pp. 577--585, 2008.

\bibitem{Michel2010}
F.~Michel and N.~Paragios,
\newblock ``{Image transport regression using mixture of experts and discrete
  markov random fields},''
\newblock in {\em Int. Symp. Biomed. Imag.}, 2010, pp. 1229--1232.

\bibitem{Avants2009}
B.~B. Avants, C.~L. Epstein, M.~Grossman, and J.~C. Gee,
\newblock ``{Symmetric Diffeomorphic Image Registration with Cross-Correlation:
  Evaluating Automated Labeling of Elderly and Neurodegenerative Brain},''
\newblock {\em Med Image Anal}, vol. 12, no. 1, pp. 26--41, 2009.

\bibitem{Pietzsch2015}
T.~Pietzsch, S.~Saalfeld, S.~Preibisch, and P.~Tomancak,
\newblock ``{BigDataViewer: visualization and processing for large image data
  sets.},''
\newblock {\em Nature Methods}, vol. 12, no. 6, pp. 481--483, may 2015.

\bibitem{Schindelin2012}
J.~Schindelin, I.~Arganda-Carreras, E.~Frise,  et~al.,
\newblock ``{Fiji: an open-source platform for biological-image analysis.},''
\newblock {\em Nature methods}, vol. 9, no. 7, pp. 676--82, jul 2012.

\bibitem{Becker2013a}
C.~Becker, R.~Rigamonti, V.~Lepetit, and P.~Fua,
\newblock ``{Supervised feature learning for curvilinear structure
  segmentation},''
\newblock in {\em MICCAI}, 2013, vol. 8149 LNCS, pp. 526--533.

\bibitem{Kothe1999}
U.~K\"{o}the,
\newblock ``{Reusable Software in Computer Vision},''
\newblock in {\em Handbook on Computer Vision and Applications}, P.~G. {B.
  J{\"{a}}hne, H. Hau{\ss}ecker}, Ed. 1999, vol.~3, Academic Press.

\end{thebibliography}

\end{document}